\documentclass{article}
\usepackage{spconf,amsmath,graphicx}

\usepackage{amsfonts}
\usepackage{microtype}
\usepackage{color}
\usepackage{soul}
\usepackage{ifthen}
\usepackage{textcomp}
\usepackage{mathtools}
\usepackage[hidelinks]{hyperref}

\def\draft{\boolean{false}}  
\newcommand\comment[1]{%
	\ifthenelse{\draft}{\textcolor{red}{#1}}{}}
\newcommand\ins[1]{%
	\ifthenelse{\draft}{\textcolor{red}{#1}}{#1}}
\newcommand\del[1]{%
	\ifthenelse{\draft}{\textcolor{red}{\st{#1}}}{}}

\title{Adversarial Semi-Supervised Audio Source Separation \\applied to Singing Voice Extraction}
\name{Daniel Stoller$^1$, Sebastian Ewert$^2$, Simon Dixon$^1$\thanks{This work was partially funded by EPSRC grants EP/L01632X/1 and EP/L019981/1 and was conducted while S.~Ewert was at Queen Mary University of London. Implementation available at \url{https://github.com/f90/AdversarialAudioSeparation}}}
\address{$^1$ Queen Mary University of London, UK  \;\;\; $^2$ Spotify\\
		\{d.stoller, s.ewert, s.e.dixon\}@qmul.ac.uk}
        
\begin{document}
\ninept

\maketitle

\begin{abstract}
The state of the art in music source separation employs neural networks trained in a supervised fashion on multi-track databases to estimate the sources from a given mixture. With only few datasets available, often extensive data augmentation is used to combat overfitting. Mixing random tracks, however, can even reduce separation performance as instruments in real music are strongly correlated. The key concept in our approach is that source estimates of an optimal separator should be indistinguishable from real source signals. Based on this idea, we drive the separator towards outputs deemed as realistic by discriminator networks that are trained to tell apart real from separator samples. This way, we can also use unpaired source and mixture recordings without the drawbacks of creating unrealistic music mixtures. Our framework is widely applicable as it does not assume a specific network architecture or number of sources. To our knowledge, this is the first adoption of adversarial training for music source separation. In a prototype experiment for singing voice separation, separation performance increases with our approach compared to purely supervised training.
\end{abstract}
\section{Introduction}
\label{sec:intro}

\begin{figure}[t]
\begin{minipage}[t]{0.48\textwidth}
  \centering
  \centerline{\includegraphics[width=7.5cm]{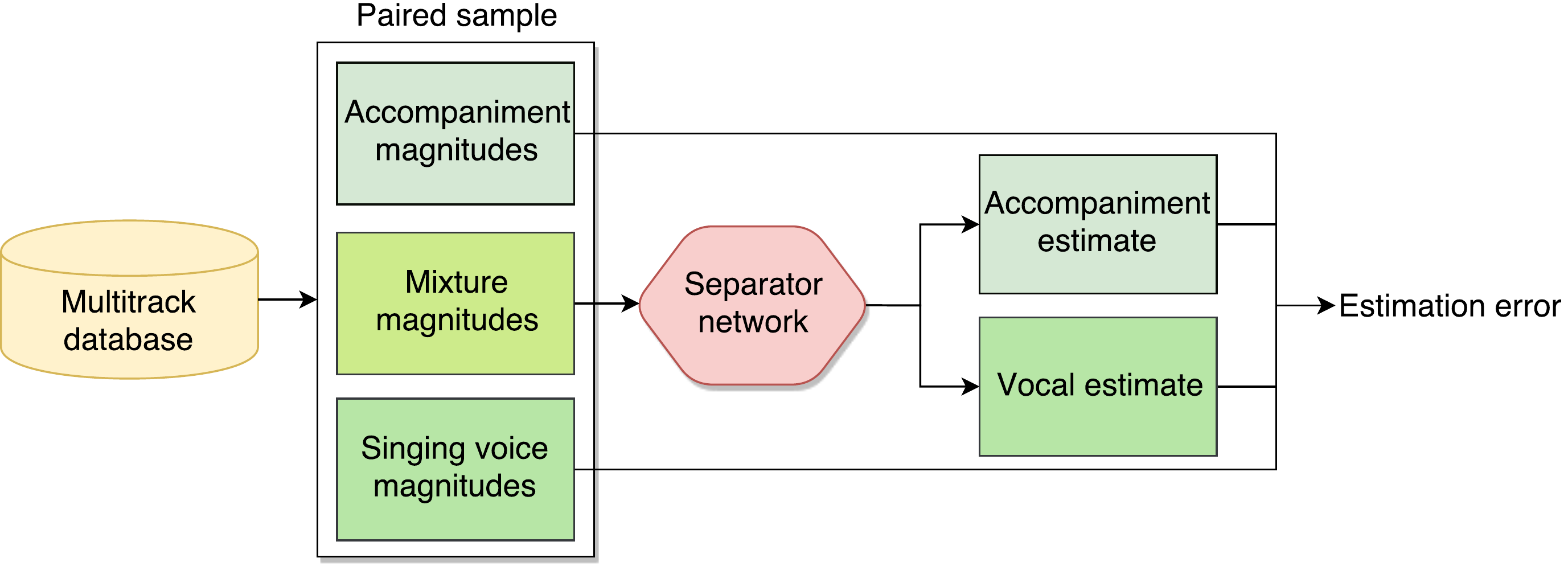}}
  \centerline{(a) Standard supervised training}\medskip
\end{minipage}
\begin{minipage}[b]{0.48\textwidth}
  \centering
  \centerline{\includegraphics[width=8.5cm]{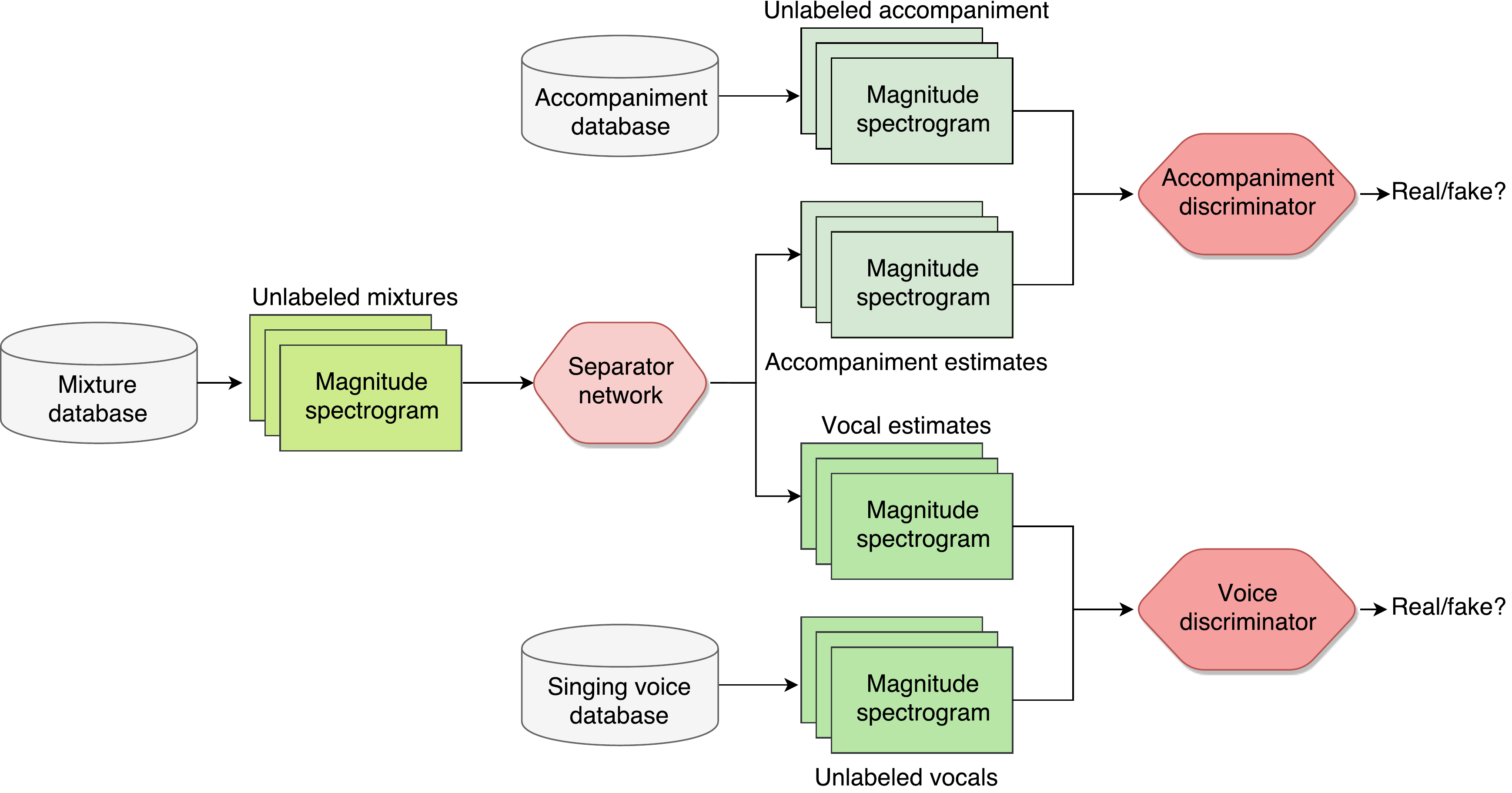}}
  \centerline{(b) Proposed adversarial training}\medskip
\end{minipage}
\vspace{-0.4cm}
\caption{Our proposed system, shown for the example of singing voice separation. In addition to a supervised loss (a), we add an unsupervised loss that drives the separator to produce a more believable output distribution for each source, as assessed by discriminators that try to distinguish real from separator samples.}
\label{fig:res}
\end{figure}

Separating instruments from music recordings is challenging as the individual sources are highly correlated in both time and frequency.
To approach such a setting, most current methods train deep networks trained in a supervised manner to directly approximate the posterior distribution over sources for a given mixture input.
Since the source estimate is compared to the target for each input, as shown in Figure~\ref{fig:res}(a), this requires paired input-output samples from multi-track recordings~--~unfortunately, publicly available datasets are rather small, which limits the overall performance.
As a result, data augmentation (and other regularisation techniques) are used to combat overfitting~\cite{Uhlich2017,Miron2017}.
However, some of the assumptions made are unrealistic: for example, randomly mixing sources implicitly assumes that sources in a music recording are independent -- the reason that music separation is so difficult, however, is exactly due to correlation between instruments.
As a result, performance can be limited since source correlations in the test set cannot be learned from the augmented training data.

With a generative approach, however, we can instead model a prior over the sources and how they interact to produce a mixture, the former of which can be learned from solo source recordings.
Separation is then an inference problem amounting to finding source estimates that explain a given mixture under the generative model.
Since modeling source priors and performing posterior inference is computationally intensive, models often have to be heavily simplified~\cite{Fevotte2007}, again limiting their performance.

In this paper, we therefore develop a novel unsupervised objective shown in Figure~\ref{fig:res}(b) that makes use of the large amount of available unlabelled music tracks as well as datasets of solo source instrument recordings, and combine it with supervised training.
This way, we can benefit from the numerical behaviour and training stability of supervised methods while also capturing the variability and richness found in large amounts of unlabelled data.
In particular, one discriminator network per source is continually trained to distinguish separator estimates made on the unlabelled music from real samples taken from the respective source dataset.
The separator aims to output more realistic sources as judged by the discriminators, in addition to minimising the supervised loss on multi-track data.

Overall, we make the following contributions:
\begin{itemize}
\item{Our theoretical framework for semi-supervised source separation can harness all available source and mixture data;}
\item{The framework can be used with any neural network architecture and type of audio representation;}
\item{Adversarially enforced source priors enable training larger models with better performance despite limited multi-track data, without regularisation such as data augmentation;}
\item{In a prototype experiment, we show performance improvements in singing voice separation over a baseline model.}
\end{itemize}

\section{Related work}
\label{sec:related_work}

\subsection{Generative approaches}
A Bayesian perspective is well suited for source separation if extensive prior knowledge about the sources is available, as it can be explicitly integrated into the model.
However, early approaches~\cite{Fevotte2007,Cemgil2007} often have to make many simplifying assumptions about the data generation process to  constrain the generative model such that the difficult problem of posterior inference is tractable.
A more recent framework provides various entry points to incorporate available prior knowledge into source models~\cite{Ozerov2012}. However, the resulting complex models do not always scale to large data due to the use of  computationally intensive inference algorithms.

For modeling singing voice, a commonly made assumption is a sparse representation in the magnitude spectrogram, while the accompaniment is of low rank and changes more slowly~\cite{Huang2012,Chan2015,Rafii2013a}.
However, as indicated by recent evaluation campaigns~\cite{Liutkus2017}, modelling more nuanced relationships (using deep networks) might be beneficial since this assumption only holds to an extent.

Non-negative matrix factorisation (NMF) is often used for separation, and can elegantly incorporate prior knowledge about sources by adapting spectral templates~\cite{Sun2013}.
However, NMF is limited in expressivity due to the assumption that spectral content can be factorised independently of time~\cite{Ewert2014}, and many spectral bases are needed to represent complex instruments.

Overall, current generative models are subject to various constraints in their structure, sacrificing separation performance to keep inference tractable, or require expert knowledge to set priors.
In contrast, we use deep networks to enforce source priors since they make minimal assumptions about the source properties and instead acquire them from data.

\subsection{Direct posterior approximation}

Many deep neural networks have been trained to directly predict sources from mixture input, from feed-forward~\cite{Nugraha2016} to convolutional~\cite{Simpson2015,Miron2017} and recurrent neural networks~\cite{Huang2014, Luo2017, Uhlich2017}.
The loss involves comparing the prediction and the correct output for each input, restricting the approach to input-output pairs from multi-track datasets.
Although these deep architectures perform well, it is not directly possible to improve them by also learning from individual source recordings, since the prior is not explicitly modelled.

Furthermore, extensive data augmentation is required~\cite{Uhlich2017, Miron2017} to combat overfitting due to the limited number of multi-track recordings.
Randomly mixing source excerpts to generate mixtures is common, but assumes that sources are temporally independent.
Since this is not the case in music pieces, correlations between sources can not be exploited by the separator.
Our unsupervised loss does not introduce such a bias since it only enforces the overall separator output to match the real source distribution for each source separately.

\section{Proposed framework}
\label{sec:format}

Our goal is to separate a mixture $m$ into $K$ sources $\mathbf{s} = (s^1, \ldots, s^K)^T$.
Here, each such sample is as an excerpt from a magnitude spectrogram -- our framework, however, is easily adapted to other input representations such as waveforms.
Overall, we assume a multi-track dataset $\mathcal{D}_{\text{m}} = \{(\mathbf{s}_1, m_1), \ldots (\mathbf{s}_M, m_M)\}$ with $M$ input-output samples is available.
Furthermore we have access to $U$ unlabelled mixture samples $\mathcal{D}_{\text{u}} = \{m^{\text{u}}_1, \ldots, m^{\text{u}}_U\}$ and a collection $\mathcal{D}_{\text{s}}^k$ of solo recordings for each source $k$.
Let $p(\mathbf{s},m)$ be the true probability of any source-mixture pair.
We assume $\mathcal{D}_{\text{m}}$ is sampled from~$p$, $\mathcal{D}_{\text{u}}$ from the marginal $p_{\text{m}}(m) = \int_{\mathbf{s}} p(\mathbf{s}, m)$, and $\mathcal{D}_{\text{s}}^k$ from the marginal of the $k$-th source $p_{\text{s}}^k(s^k) = \int_{\{s^1,\ldots,s^K,m\}\setminus\{s^k\}} p(\mathbf{s},m)$.

We aim to train a deterministic separation function $f_{\phi}$ parametrised by a deep neural network with all available data so that $q_{\phi}(\mathbf{s}|m) = \delta(f_{\phi}(m) - \mathbf{s})$ approximates the real posterior $p(\mathbf{s}|m)$.
Current approaches usually use the mean squared error between estimates and targets for each input
\begin{equation} 
L_{\text{s}} = \frac{1}{M} \sum_{i=1}^{M} || f_{\phi}(m_i) - \mathbf{s}_i ||_2
\end{equation}
as a loss function on multi-track data.
However, this loss function does not include the unlabelled data and is minimised when predicting the posterior mean, which is an unlikely estimate itself and often corresponds to a blurred average of real posterior modes.

We derive an unsupervised loss without these issues.
An optimal separator $q_{\phi}$ would estimate the real posterior perfectly and thus fulfil $q_{\phi}(\mathbf{s}|m) = p(\mathbf{s}|m)$ for all possible $m$.
In this case, it follows that the marginal separator output $\prescript{\text{out}}{}{q}_{\phi}(\mathbf{s}) = E_{m \sim p_{\text{m}}}\ q_{\phi}(\mathbf{s}|m)$ would be equal to the true source marginal $p_{\text{s}}(\mathbf{s}) = E_{m \sim p_{\text{m}}}\ p(\mathbf{s}|m)$.
If the joint distributions $\prescript{\text{out}}{}{q}_{\phi}$ and $p_{\text{s}}$ are the same, then so are their source marginals -- with $\prescript{\text{out}}{}{q}^{k}_{\phi}(s^k) = \int_{\{s^1,\ldots,s^K\}\setminus\{s^k\}}\ \prescript{\text{out}}{}{q}_{\phi}(\mathbf{s})$, this means
\begin{equation}
\label{eq:output_condition}
\prescript{\text{out}}{}{q}^{k}_{\phi} = p^k_{\text{s}},\ \forall\ k=1,\ldots,K.
\end{equation}
The above distribution equalities are thus necessary, but not sufficient conditions for an optimal separator.

To approximately fulfil the equalities in~\eqref{eq:output_condition}, we can define a divergence $D[\prescript{\text{out}}{}{q}^k_{\phi} || p^k_{\text{s}}]$ with $D[q||p] \geq 0$ and ${D[q||p] = 0} \Leftrightarrow {q = p}$ between the two distributions for each source $k$ to formulate an unsupervised loss we aim to minimise
\begin{equation}
L_{\text{u}} = \sum_{k=1}^K D[\prescript{\text{out}}{}{q}^k_{\phi} || p^k_{\text{s}}].
\end{equation}
This loss allows using our unlabelled data since we compare $K$ pairs of source distributions instead of individual samples.
We approximate $p^k_{\text{s}}$ and $\prescript{\text{out}}{}{q}^k_{\phi}$ using batches of samples from the source dataset $\mathcal{D}_{\text{s}}^k$ and the unlabelled mixture dataset $\mathcal{D}_{\text{u}}$, respectively. 

\subsection{Measurement and choice of divergence}

To determine $L_{\text{u}}$, we need to choose a divergence D and a method to re-estimate it after each separator training step since $\prescript{\text{out}}{}{q}_{\phi}$ changes.
One possibility is to choose the \textit{Jenson-Shannon} (JS) or \textit{Kullback-Leibler} (KL) divergence as $D$ and use a discriminator network $D_{\theta_k}$ for each source $k$ that distinguishes separator from real samples to estimate each divergence -- note that this is a non-trivial result, see~\cite{Goodfellow2014} for details.
With this choice, our unsupervised loss is similar to generative adversarial networks (GANs)~\cite{Goodfellow2014}, but we use one discriminator for each source instead of only one, and our ``generator" $q_{\phi}$ receives mixtures as input instead of random noise.

GANs are known to be unstable~\cite{Arjovsky2017a} since KL and JS divergences are maximised for pairs of distributions that do not overlap, which likely happens in our setting since we use finite sets of samples for $\prescript{\text{out}}{}{q}^{k}_{\phi}$ and $p^k_{\text{s}}$ so they become dirac-like.
As a result, the gradients for the separator can vanish, or become arbitrarily large near the discriminator's decision boundary, which can destabilise training.

For a stable optimisation, we consider more well-behaved divergences such as the \textit{Wasserstein distance} $W_P$.
As discussed in~\cite{Gulrajani2017}, the gradient of $W_P$ with respect to the separator output has a bounded norm since a regularising term $L_{\text{grad}}$ is applied on the discriminator networks.
Thus we expect separator training to be more stable since the gradient applied to its output does not vanish or explode.

Following the improved Wasserstein GAN algorithm~\cite{Gulrajani2017}, we use one discriminator network $D_{\theta_k}$ for each source $k$ to approximate $W_P[\prescript{\text{out}}{}{q}^k_{\phi}||p^k_{\text{s}}]$.
We modify the gradient penalty for the discriminator to be one-sided since it enabled faster convergence in our experiments:
\begin{equation}
L_{\text{grad}} = E_{x \sim \hat{p}} \max(||\nabla_{x} D_{\theta_k}(x)||_2 - 1, 0)^2
\end{equation}
Sampling from $\hat{p}$ involves randomly interpolating between pairs of points sampled from the data and the generator distribution.
We use the Wasserstein distance for all following experiments due to instabilities we observed in initial tests using the KL and JS divergences.

\subsection{Additive penalty}
\label{sec:additive_penalty}

For all unlabelled mixtures $m^{\text{u}}_i$, we also aim to ensure that source estimates add up to the mixture, so that $\sum_{k=1}^K f_{\phi}(m^{\text{u}}_i)_k \approx m^{\text{u}}_i$. 
If we know this additive property holds exactly in the true distribution~$p$, and $q_{\phi}$ is constrained to only output estimates satisfying this constraint, no  additional loss is needed.
When using spectrograms, this relationship is only approximate due to phase interference, so we do not constrain the network output $f_{\phi}(m)$ while minimising the loss
\begin{equation}
L_{\text{add}} = \frac{1}{U} \sum_{i=1}^{U} || \sum_{k=1}^K f_{\phi}(m^{\text{u}}_i)_k - m^{\text{u}}_i ||_2
\end{equation}
which is equivalent to maximising $\sum_{i=1}^U \log p(m^{\text{u}}_i|f_{\theta}(m^{\text{u}}_i))$ as likelihood term when assuming $p(m|\mathbf{s})$ is an isotropic Gaussian $\mathcal{N}(m|\sum_{k=1}^K s_k ; \sigma^2 \mathbf{I})$.
Since we are considering aggregate priors with $L_{\text{u}}$ and a likelihood term $p(m|\mathbf{s})$ with $L_{\text{add}}$, our overall unsupervised training exhibits similarities to variational inference with $q_{\phi}$ as inference network that aims to approximate the posterior of the generative model $p(\mathbf{s},m) = p(m|\mathbf{s}) \prod_{k=1}^K p^k_{\text{s}}(s^k)$.

\subsection{Semi-supervised loss}

Overall, we minimise the total separator loss $L = L_{\text{s}} + \alpha L_{\text{u}} + \beta L_{\text{add}}$ by stochastic gradient descent using a batch of multi-track samples from $\mathcal{D}_{\text{m}}$ and a batch of unlabelled mixtures from $\mathcal{D}_{\text{u}}$.
After each separator update, we take $N_{\text{disc}}$ gradient steps for each discriminator $D_{\theta_k}$ to estimate the divergence $W_P[\prescript{\text{out}}{}{q}^k_{\phi}||p^k_{\text{s}}]$ using one shared batch of unlabelled mixtures to generate source estimates and one batch from the respective source dataset $\mathcal{D}^k_{\text{s}}$.
The scalars $\alpha$ and $\beta$ are weights for the loss terms and constitute hyper-parameters.

\section{Singing voice separation experiment}
\label{sec:experiment}

\subsection{Initial considerations}
Since the accompaniment in singing voice separation has a very complex distribution, it is harder for the discriminator to estimate the divergence $D$ than it is for the singing voice.
Therefore, we conducted one experiment without the accompaniment discriminator.
Note that after removing a divergence term from $L_{\text{u}}$ it still represents a necessary, albeit weaker, condition for an optimal separator, thus retaining the global minima of the original loss $L_{\text{u}}$.
In practice however, we may not find a global minimum, which could bias solutions towards favouring vocal over accompaniment quality.

\subsection{Datasets}

We use the training partition of the DSD100~\cite{Liutkus2017} database as our supervised training set $\mathcal{D}_{\text{m}}$.
We split the multi-track databases iKala~\cite{Chan2015}, MedleyDB~\cite{Bittner2014} and CCMixter~\cite{Liutkus2015} into thirds, and use one third of tracks from each database to form the unlabelled dataset $\mathcal{D}_{\text{u}}$ and the source datasets $\mathcal{D}^k_{\text{s}}$ needed for semi-supervised training.
Our validation and test set is each built by taking another third of tracks from iKala, MedleyDB, and CCMixter, in addition to 25 tracks from the test partition of DSD100.
The supervised and unsupervised sets have a different sampling bias to enable testing the regularization effect of our semi-supervised approach more directly.
We use multi-track data for the unsupervised set despite their known pairing to eliminate dataset bias as a confounding factor, ensuring differences between the separator output $\prescript{\text{out}}{}{q}_{\phi}$ and the source dataset distributions stem from the separator.
Large databases such as DAMP~\cite{Smith2013} could be used as unsupervised data assuming they are sufficiently similar to multi-track stems -- however, we only aimed to provide a first proof-of-concept in this paper and thus did not include such datasets.

\subsection{Experimental setup}
\subsubsection{Preprocessing}

\begin{figure*}[t!]
\scalebox{0.9}{
\footnotesize
\begin{tabular}{|c|ccc|ccc|ccc|ccc|ccc|}
\hline
& \multicolumn{3}{|c|}{Test set} & \multicolumn{3}{|c|}{DSD100} & \multicolumn{3}{|c|}{MedleyDB} & \multicolumn{3}{|c|}{CCMixter} & \multicolumn{3}{|c|}{iKala} \\
& Baseline & V & VA & Baseline & V & VA & Baseline & V & VA & Baseline & V & VA & Baseline & V & VA \\
\hline
SDR Inst. & 8.09 & \textbf{8.89} & 8.55 & \textbf{11.11} & 10.75 & 10.76 & 9.40 & 9.60 & \textbf{9.65} & 10.65 & \textbf{11.09} & 10.89 & 6.34 & \textbf{7.71} & 7.13 \\
SDR V. & 6.80 & 7.28 & \textbf{7.47} & \textbf{3.74} & 3.17 & 3.54 & 2.48 & 2.43 & \textbf{3.00} & 3.25 & 3.52 & \textbf{3.70} & 9.50 & 10.47 & \textbf{10.52} \\
SIR Inst. & 12.03 & 12.58 & \textbf{12.67} & \textbf{14.46} & 13.56 & 13.86 & 12.18 & 12.07 & \textbf{12.74} & 15.99 & 15.49 & \textbf{16.08} & 10.42 & \textbf{11.79} & 11.57 \\
SIR V. & 13.72 & 14.00 & \textbf{14.45} & \textbf{10.03} & 9.92 & 10.49 & 9.40 & 9.21 & \textbf{9.48} & 8.39 & 8.94 & \textbf{9.35} & 16.98 & 17.44 & \textbf{17.90}\\
SAR Inst. & 11.27 & \textbf{12.05} & 11.40 & 14.20 & \textbf{14.60} & 14.10 & 13.94 & \textbf{14.23} & 13.45 & 12.84 & \textbf{13.69} & 13.24 &  9.43 & \textbf{10.42} & 9.70\\
SAR V. & 8.54 & 9.00 & \textbf{9.04} & \textbf{5.50} & 4.84 & 5.12 & 4.71 & 4.69 & 5.20 & \textbf{6.43} & 6.17 & 6.17 & 10.81 & \textbf{11.83} & 11.73\\
\hline
\end{tabular}}
\caption{Mean test set performance comparison on the test set (22 instrumental tracks excluded, mono, 8 KHz sampling rate) and subsets using the supervised baseline, using a vocal discriminator (V) and using both vocal and accompaniment discriminators (VA)}
\label{tab:results}
\end{figure*}

The audio input is converted to mono and downsampled to $8$ KHz to reduce dimensionality, before the magnitude spectrogram is computed from a 512-point FFT with 50\% overlap, and normalized by ${x \rightarrow }\log(1+x)$.
For the unsupervised dataset, we multiply the magnitudes by a factor uniformly drawn from the interval $[0.2, 1.2]$ to induce invariance to loudness differences in the source discriminators.
We randomly draw $64$ spectrogram excerpts for each batch.

\subsubsection{Separator architecture}
The separator architecture follows the U-Net~\cite{Ronneberger2015,Jansson2017} closely and uses $3 \times 3$ convolutional filters.
After a first convolutional layer with $16$ filters, $4$ downsampling layers perform max-pooling by a factor of $2$ followed by a convolution with twice as many filters as the last layer.
The $4$ upsampling layers perform transposed convolution with a stride of $2$, crop and concatenate the feature map from the respective downsampling layer before applying another transposed convolution.

The last feature map is concatenated with the mixture input so it can be used as basis for the output, before $K$ separate 1x1 convolutions are applied, one for each source.
After applying ReLU activations, the $K$ feature maps form the $K$ log-normalised source estimates, which are directly input to the discriminators.
To generate the final source signals, we use an inverse STFT using the phase from the mixture input.
Since the U-Net requires additional context to make predictions at the centre of its input, we use valid convolutions, add temporal context to the input and zero-pad along the frequency axis. We input $158$ $350$-dimensional time frames to retrieve $66$ $256$-dimensional time frames as output.

\subsubsection{Discriminator architecture}
The source discriminators receive log-normalised magnitude spectrograms and follow the DCGAN architecture~\cite{Radford2015} with Leaky ReLU activations and $32$ convolutions in the first layer, with zero-padding in time and frequency. 
Since the input samples have more frequency bins than time frames, we use two convolutional layers with $4 \times 2$ filters and $2 \times 1$ stride after the first four layers to detect relationships across frequency bands, before computing 32 dense activations and finally a single linear output.

\subsubsection{Training procedure}
We train the separator and discriminators on an NVIDIA GTX1080 using the ADAM optimiser with a learning rate of $5 \cdot 10^{-5}$.
Training is stopped if validation performance does not increase after more than six epochs, with 1000 separator update steps in each epoch.
Finally, the model with the best validation loss is selected.

A baseline model is trained using only the supervised loss in the log-normalised magnitude space.
Then, we train a network with the same architecture using our semi-supervised approach with $\alpha = 0.01$, but without accompaniment discriminator.
Finally, we use both discriminators and a lower $\alpha=0.001$.
Each time, we set $\beta = \alpha$.
We use low values for $\alpha$ since in initial tests the loss occasionally plateaued during training, likely due to local minima in the unsupervised loss.
Discriminators are trained for $N_{\text{disc}} = 5$ iterations per separator update to re-estimate the respective Wasserstein distance.

\subsection{Evaluation}
\subsubsection{Quantitative results}

For evaluation, we calculate the track-wise (normalised) SDR, SIR, and SAR metrics~\cite{Vincent2006}, with mono estimates and target signals sampled at 8 KHz. Table~\ref{tab:results} shows averages over the test set and its subsets containing only tracks from a specific data source.
On the full test set, the purely supervised method (`Baseline') is consistently improved upon across every metric by our method, both with and without an accompaniment discriminator.
The baseline method is only better on the DSD100 subset, likely because the supervised set contains only DSD100 training samples, which can be viewed as overfitting.
On all other datasets our method yields improvements, especially on the iKala dataset, showing we can train the separator on these samples despite not knowing their input-output pairings.
Therefore our unsupervised loss can be a surrogate for the supervised loss enabling learning from unlabelled mixture and source datasets.

\subsubsection{Qualitative analysis}
\begin{figure}[t]
\begin{minipage}[l]{0.48\linewidth}
  \centering
  \centerline{\includegraphics[width=\textwidth]{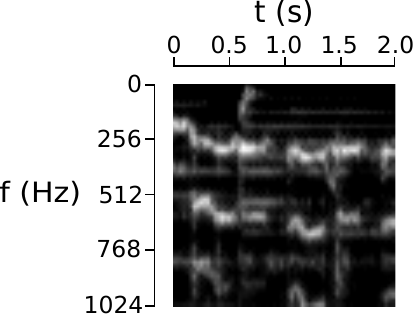}}
  \centerline{(a) Separator estimate $x$}\medskip
\end{minipage}
\begin{minipage}[r]{0.48\linewidth}
  \centering
  \centerline{\includegraphics[width=\textwidth]{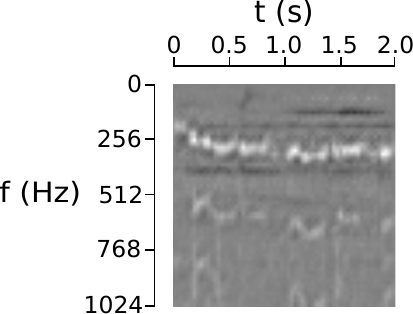}}
  \centerline{(b) $\nabla_x D(x)$}\medskip
\end{minipage}
\caption{(a) A separator voice estimate $x$. (b) Gradients of the voice discriminator output with respect to the input $x$. Only the lower frequency range is shown.}
\label{fig:discriminator}
\end{figure}

For an intuition about the discriminator's behaviour, Figure~\ref{fig:discriminator}(a) shows an exemplary vocal estimate from the separator during training, where white denotes high energy.
Next to a strong singing voice with vibrato, accompaniment interference is visible as straight, horizontal lines.
The discriminator was successfully trained to output large values for real and low values for separator samples:
The gradient with respect to the input is positive (shown in white) for vocal parts and negative (shown in black) for the accompaniment artefacts.
Therefore the separator is encouraged to attenuate the accompaniment and amplify the voice content to make the voice output more realistic.

\section{Conclusion}
\label{sec:conclusion}

We presented a semi-supervised framework for audio source separation.
In addition to supervised training on multi-track data, we introduce an unsupervised loss on unlabelled mixtures driving the separator to minimise a divergence between its output distribution and the real source distribution, for each source.
The divergence for each source is estimated by its own discriminator continually trained to distinguish real source samples from separator predictions. 
Our framework is scalable since it can acquire complex source priors from large amounts of unlabelled data while making only few assumptions about the source characteristics.

For singing voice separation, we show an increase in performance compared to purely supervised training.
However, performance can also be reduced if the unlabelled data is too scarce or does not come from the same distribution as the test set.  
Therefore, we used multi-track datasets as our unlabelled data in our initial experiment to avoid this confounding factor, but datasets such as DAMP~\cite{Smith2013} could be included if the dataset bias is slight or can be controlled.

Future work could involve applying our framework to multi-instrument separation due to the highly structured priors for many sources.
Our semi-supervised approach also allows training larger separator models that would not generalise sufficiently when trained on multi-track data alone.
Finally, discriminator architectures could be adapted to better distinguish separator from real samples and to be less sensitive to the inherent source variability.

\bibliographystyle{IEEEbib}
\bibliography{refs}

\end{document}